\documentclass[letterpaper, 10 pt, conference]{ieeeconf}  

\IEEEoverridecommandlockouts                              

\usepackage{graphics} 
\usepackage{graphicx}
\usepackage{amsmath,amssymb}
\usepackage{hyperref}

\title{MDPO: Multi-constraint Dexterous Policy Optimization with\\ Reinforcement Learning}

\author{
Ao Zhou$^{1,2}$, Zongyou Yu$^{1,2}$, Ke Wu$^{3,4}$, Tao Xu$^{1,2}$, Junqiao Zhao$^{1}$,\\ Wenchao Ding$^{3,4}$, Alois Knoll$^{5}$, Guang Chen$^{1,2}$*
\thanks{$^{*}$ Corresponding author: Guang Chen, Email: guangchen@tongji.edu.cn}
\thanks{Author's Affiliation: $^1$Tongji University, Shanghai, China; $^2$Shanghai Innovation Institute, Shanghai, China; $^3$Fudan University, Shanghai, China; $^4$TARS Robotics, Shanghai, China; $^5$Technical University of Munich, Munich, Germany.}
}

\begin{document}

\maketitle
\thispagestyle{empty}
\pagestyle{empty}


\begin{abstract}
Dexterous manipulation requires high-dimensional control policies to accomplish complex contact-rich tasks while simultaneously satisfying multiple safety requirements, such as joint limits, unsafe finger configurations, arm collisions, and unstable object interactions. Constrained reinforcement learning (RL) provides a principled framework for incorporating these requirements into dexterous manipulation policy learning. However, existing constrained RL methods for dexterous manipulation primarily focus on cumulative cost constraints and struggle to simultaneously handle multiple heterogeneous constraints during dexterous policy learning. In this work, we propose \textbf{M}ulti-constraint \textbf{D}exterous \textbf{P}olicy \textbf{O}ptimization~(MDPO) for safe dexterous manipulation. We first formulate dexterous manipulation as a multi-constraint policy optimization problem involving both cumulative cost constraints and state-wise cost constraints. Based on this formulation, we integrate multi-constraint projection into the dexterous manipulation policy optimization process. Specifically, we employ constraint-specific policy projections to map the intermediate policy that maximizes task performance onto the feasible regions defined by the cumulative cost constraint and the state-wise cost constraint. Extensive experiments on ReDMan, including safe joint control, finger safety, and grasping tasks, demonstrate that MDPO achieves superior constraint satisfaction while maintaining competitive task performance compared with the baselines. Our code is available at this \href{https://github.com/sii-pilab}{link}.
\end{abstract}
\section{INTRODUCTION}
Dexterous manipulation is a fundamental capability for embodied robots, enabling them to perform complex tasks that require high-dimensional coordination, precise contact control, and flexible interactions with objects~\cite{nagabandi2020deep}. Recent advances in reinforcement learning (RL) have demonstrated considerable potential for learning dexterous manipulation policies directly through interactions with the environment~\cite{yuan2025cross}. Unlike conventional model-based approaches that rely on accurate dynamics models and manually designed controllers~\cite{hess2025sampling}, RL provides a data-driven paradigm that optimizes manipulation strategies directly from environmental feedback~\cite{ye2023reinforcement}. This characteristic makes RL particularly attractive for high-dimensional dexterous manipulation tasks involving complex contact dynamics and nonlinear interactions. However, applying RL to real-world dexterous manipulation remains challenging due to the presence of multiple physical cost constraints. For example, excessive joint motions and repeated unsafe interactions should remain within tolerable limits throughout the manipulation process, whereas critical unsafe states, such as severe violations of joint limits or hazardous finger and arm configurations, should be strictly avoided~\cite{geng2025redman}. Therefore, simultaneously considering dexterous manipulation performance, cumulative safety constraints, and state-wise safety constraints is crucial for achieving reliable dexterous manipulation.

Constrained RL provides a principled framework for incorporating safety requirements into policy optimization through Constrained Markov Decision Processes (CMDP)~\cite{altman2021constrained}. Although constrained RL has improved the safety of dexterous manipulation policies~\cite{geng2025redman}, existing methods mainly focus on a single cumulative cost constraint and rarely consider the simultaneous regulation of cumulative and state-wise safety requirements~\cite{zhao2023state}. Conventional penalty or Lagrangian approaches handle constraints by introducing manually designed or adaptively learned coefficients~\cite{wu2024off,zhang2022penalized}. However, extending such methods to heterogeneous constraints requires multiple penalty factors, increasing optimization difficulty and potentially leading to unstable or overly conservative policy updates. Projection-based methods provide an alternative by mapping updated policies back into feasible constraint regions~\cite{lin2024projection,yang2022constrained,yang2020projection}, but are still primarily developed for the single constraint settings. Moreover, cumulative and state-wise constraints exhibit different temporal characteristics and may interact during dexterous manipulation: improving task performance by relaxing cumulative safety can increase instantaneous safety risks. Therefore, an effective multi-constraint RL algorithm is needed to jointly regulate long-term cumulative risks and state-wise safety violations while preserving manipulation performance.

In this work, we propose \textbf{M}ulti-constraint \textbf{D}exterous \textbf{P}olicy \textbf{O}ptimization~(MDPO), a RL algorithm for multi-constraint safe dexterous manipulation. We first formulate dexterous manipulation as a multi-constraint policy optimization problem involving both cumulative cost constraints and state-wise cost constraints, allowing different safety requirements to be explicitly represented according to their physical characteristics. Based on this formulation, MDPO performs a regularized policy improvement step to obtain an intermediate policy that improves manipulation performance. The intermediate policy is then projected back into the joint safe feasible region defined by both cumulative and state-wise cost constraints. Specifically, MDPO simultaneously regulates long-term cumulative safety violations over an episode and state-wise safety-critical behaviors during individual interactions, ensuring that the updated policy satisfies both types of safety requirements. Finally, we evaluate the proposed method on the ReDMan~\cite{geng2025redman} benchmark, which contains diverse dexterous manipulation scenarios with challenging physical constraints. Experiments on safe joint control, finger-level safety, and grasping tasks demonstrate that MDPO consistently achieves better constraint satisfaction and competitive manipulation performance compared with baselines. The main contributions of this work are summarized as follows:
\begin{itemize}
\item To the best of our knowledge, we are the first to formulate dexterous manipulation as a multi-constraint RL problem, enabling simultaneous optimization of manipulation objectives, cumulative cost constraints, and state-wise cost constraints.
\item The intermediate policy is then projected back into the joint safe feasible region defined by both cumulative and state-wise cost constraints. Specifically, MDPO simultaneously regulates long-term cumulative safety violations over an episode and state-wise safety-critical behaviors during individual interactions, ensuring that the updated policy satisfies both types of safety requirements.
\item Extensive experiments on the ReDMan benchmark demonstrate that MDPO achieves superior constraint satisfaction and competitive manipulation performance across various dexterous manipulation tasks.
\end{itemize}
\section{RELATED WORK}
In this section, we review the literature most relevant to our work, including RL for dexterous manipulation and constrained RL for safe policy optimization.
\subsection{RL for Dexterous Manipulation}
Dexterous manipulation requires coordinated control of high-dimensional joints and contacts for complex object interactions~\cite{yang2024anyrotate}. RL has become an effective paradigm for learning such skills through direct interaction with environments~\cite{yuan2025cross}. Early studies demonstrated successful policy learning in simulation~\cite{rajeswaran2017learning} and transfer to physical robots through improved exploration and sim-to-real techniques~\cite{andrychowicz2020learning}. Recent advances have further enabled challenging tasks including grasping, object reorientation~\cite{mandi2025dexmachina}, and in-hand manipulation~\cite{khandate2023sampling}. However, most methods still emphasize task success and manipulation performance~\cite{ze2023h}, while safety is typically handled through reward penalties~\cite{hu2023causal} or early termination~\cite{khandate2023sampling}. ReDMan~\cite{geng2025redman} explicitly introduces safety costs for dexterous manipulation, but mainly considers a single cumulative constraint. Consequently, existing methods remain insufficient for manipulation scenarios involving multiple heterogeneous safety requirements~\cite{gu2025safe}.

\subsection{Constrained RL}
Constrained RL incorporates safety requirements into policy optimization through CMDP formulations~\cite{altman2021constrained}. A widely used approach is Lagrangian optimization~\cite{stooke2020responsive}, which introduces adaptive multipliers to balance reward maximization and constraint satisfaction. Other representative directions include interior-point optimization~\cite{liu2020ipo}, CVaR-based formulations~\cite{chow2015risk}, penalty-based methods~\cite{zhang2022penalized}, state augmentation~\cite{sootla2022saute}, and world-model-based safe RL~\cite{huang2024safedreamer}. Projection-based methods~\cite{yang2022constrained} instead decouple performance improvement from safety correction by projecting an intermediate policy back into the feasible region. Primal-dual approaches provide another important paradigm by jointly optimizing policies and dual variables, with representative methods including PDO~\cite{chow2018risk} and CVPO~\cite{liu2022constrained}. Despite these advances, most existing methods are developed primarily for cumulative constraints, motivating more effective optimization for heterogeneous safety requirements.

Although constrained RL has achieved considerable success in safety-critical control, existing safe dexterous manipulation studies remain largely limited to single cumulative cost constraints. A straightforward extension to multiple safety requirements is to introduce additional Lagrange multipliers or penalty coefficients, which increases the number of constraint-related hyperparameters and may lead to unstable or overly conservative policy optimization. In particular, cumulative and state-wise safety constraints in dexterous manipulation are inherently coupled: relaxing the cumulative cost constraint may improve task performance, but can simultaneously increase the likelihood of violating the state-wise cost constraint. These limitations highlight the need for more effective multi-constraint formulations for safe dexterous manipulation.
\section{PRELIMINARIES}
RL is commonly formulated as the Markov Decision Process (MDP) defined by the tuple $\mathcal{M}=(\mathcal{S},\mathcal{A},P,r,\rho_0,\gamma)$, where $\mathcal{S}$ and $\mathcal{A}$ denote the state and action spaces, respectively, $P(s_{t+1}|s_t,a_t)$ is the transition probability, $r:\mathcal{S}\times\mathcal{A}\rightarrow \mathbb{R}$ denotes the reward function, $\rho_0$ is the initial-state distribution, and $\gamma\in(0,1]$ is the discount factor. A parameterized policy $\pi_\theta(a_t|s_t)$ determines the action distribution conditioned on the current state. A trajectory generated by $\pi_\theta$ is denoted as $\tau=\{s_0,a_0,s_1,a_1,\ldots\}\sim\pi$, where $s_0\sim\rho_0$, $a_t\sim\pi_\theta(\cdot|s_t)$, and $s_{t+1}\sim P(\cdot|s_t,a_t)$. The objective of RL is to maximize the rewards:
\begin{equation} 
    \label{equ3_1}
    \begin{aligned}     
     \pi^{*}\!=\!\arg \max_{\pi}\mathbb{E}_{\tau\sim\pi}\!\left[\sum_{t=0}^{\infty}\gamma^{t}r_{t}\!\right]\!
    \end{aligned}.
\end{equation}
In dexterous manipulation, the reward generally reflects task objectives such as successful grasping, object manipulation, pose tracking, or task completion.

Standard RL only maximizes task rewards and does not explicitly restrict unsafe behaviors. Constrained RL extends the MDP to the CMDP by introducing additional cost functions to characterize safety violations. The CMDP is defined by $\mathcal{M}_c=(\mathcal{S},\mathcal{A},P,r,c,\rho_0,\gamma)$, where $c:\mathcal{S}\times\mathcal{A}\rightarrow \mathbb{C}$ denotes the cost function. We use shorthand $c_t=c(s_t,a_t)$ for simplicity. Given a predefined cost threshold $\bar{c}$, constrained RL aims to maximize task performance while satisfying the safety constraints:
\begin{equation} 
    \label{equ3_2}
    \begin{aligned}     
    \!\!\pi^{*}\!=\!\arg \max_{\pi}\mathbb{E}_{\tau\sim\pi}\!\left[\sum_{t=0}^{\infty}\gamma^{t}r_{t}\!\right]\!,\text{s.t.}\mathbb{E}_{\tau\sim\pi}\!\left[\sum_{t=0}^{\infty}\gamma^{t}c_{t}\!\right]\!\leq\! \bar{c}
    \end{aligned},
\end{equation}
Conventional CMDP formulations mainly characterize safety through cumulative costs. However, dexterous manipulation involves safety requirements with different temporal characteristics. Some violations can be tolerated occasionally as long as their cumulative occurrence remains bounded, whereas safety-critical configurations should be restricted at individual interaction steps. Following this consideration, we formulate safe dexterous manipulation with two types of safety constraints: cumulative cost constraints and state-wise cost constraints. Let $\bar{c}$ denote the cumulative cost threshold and its constraint is defined as $\mathbb{E}_{\tau\sim\pi}\!\left[\sum_{t=0}^{\infty}\gamma^{t}c_{t}\right]\!\leq\! \bar{c}$. We further introduce state-wise cost threshold $\bar{D}$ and its constraint is defined as $\forall t\geq 0,~\mathbb{E}_{s_{t}\sim p(\cdot|s_t,a_t), a_{t}\sim \pi(\cdot|s_t)}[c_{t}] \leq \bar{D}$. Accordingly, the multi-constraint safe dexterous manipulation objective is formulated as:
\begin{equation} 
    \label{equ3_3}
    \begin{aligned}     
    &\qquad\qquad\pi^{*}\!=\!\arg \max_{\pi}\mathbb{E}_{\tau\sim\pi}\!\left[\sum_{t=0}^{\infty}\gamma^{t}r_{t}\!\right]\!,\\
    &\qquad\qquad\:\:\: \text{s.t.}\:\:\mathbb{E}_{\tau\sim\pi}\!\left[\sum_{t=0}^{\infty}\gamma^{t}c_{t}\!\right]\!\leq\! \bar{c},\\
    &\:\:\:\quad\forall t\geq 0,~\mathbb{E}_{s_{t}\sim p(\cdot|s_t,a_t), a_{t}\sim \pi(\cdot|s_t)}[c_{t}] \leq \bar{D}.\\
    \end{aligned}
\end{equation}
For subsequent policy optimization, we define the feasible region induced by cumulative cost constraints as:
\begin{equation} 
    \label{equ3_4}
    \begin{aligned}     
    \Pi_{\mathrm{cum}}=\left\{\pi : \mathbb{E}_{\tau\sim\pi}\!\left[\sum_{t=0}^{\infty}\gamma^{t}c_{t}\!\right]\!\leq\! \bar{c} \right\},
    \end{aligned}
\end{equation}
and the feasible region associated with state-wise cost constraints as:
\begin{equation} 
    \label{equ3_5}
    \begin{aligned}     
    \Pi_{\mathrm{state}}=\left\{\pi : \mathbb{E}_{s_{t}\sim p(\cdot|s_t,a_t), a_{t}\sim \pi(\cdot|s_t)}[c_{t}] \leq \bar{D},~\forall t\geq 0 \right\},
    \end{aligned}
\end{equation}
Hence, the multi-constraint safe policy optimization objective can be equivalently expressed as:
\begin{equation} 
    \label{equ3_6}
    \begin{aligned} 
    \pi^{*}=\underset{\pi \in \Pi_{\mathrm{cum}} \cap \Pi_{\mathrm{state}}}{\arg\max}\;\mathbb{E}_{\tau \sim\pi}\left[\sum_{t=0}^{\infty} \gamma^{t} r_t\right].
    \end{aligned}
\end{equation}
The cumulative cost constraints regulate long-term safety violations throughout manipulation, whereas the state-wise cost constraints restrict safety-critical behaviors at individual time steps. These two types of constraints are inherently coupled in dexterous manipulation: relaxing cumulative safety constraints may improve task performance, but can simultaneously increase the likelihood of violating state-wise safety requirements. Therefore, jointly optimizing manipulation performance while satisfying both types of safety constraints constitutes the multi-constraint dexterous policy optimization problem considered in this work.
\section{METHODOLOGY}
In this section, we present Multi-constraint Dexterous Policy Optimization (MDPO) for safe dexterous manipulation under cumulative and state-wise cost constraints. We first reformulate the state-wise constraint into a tractable trajectory-level form. We then introduce the two-stage optimization framework of MDPO, consisting of policy performance improvement and joint multi-constraint policy projection. Finally, we present a sample-based first-order implementation for practical policy optimization.

\subsection{State-wise Constraint Reformulation}
The dexterous manipulation policy is simultaneously subject to a cumulative cost constraint and a state-wise cost constraint based on the same safety cost function $c$. The cumulative cost constraint can be directly evaluated over an entire trajectory as $\mathbb{E}_{\tau\sim\pi}\!\left[\sum_{t=0}^{\infty}\gamma^{t}c_{t}\right]\!\leq\! \bar{c}$. In contrast, the state-wise cost constraint requires the safety cost at each interaction step to remain below a predefined threshold $\forall t\geq 0,~\mathbb{E}_{s_{t}\sim p(\cdot|s_t,a_t), a_{t}\sim \pi(\cdot|s_t)}[c_{t}] \leq \bar{D}$. Such a state-wise constraint does not naturally take the form of a trajectory-level cumulative objective, making it difficult to directly incorporate into conventional policy optimization. To address this issue, we transform the state-wise safety requirement into an accumulated form through the state-wise cost increment. This formulation follows the maximum state-wise cost transformation used in the provided multi-constraint formulation. $M_t$ is introduced to record the maximum state-wise cost encountered before time step before time step $t$. The state-wise cost increment is defined as $D(\hat{s}_t, a_t) = \max \left\{ c(s_t, a_t) - M_{t}, \; 0 \right\}$ where $\hat{s}_t=(s_t,M_{t})$ denotes the augmented state. The maximum state-wise cost can be recursively updated as $M_{t+1}=M_{t}+D(\hat{s}_t, a_t)$, $M_0 = 0$. According to $D(\hat{s}_t, a_t) = \max \left\{ c(s_t, a_t) - M_{t}, \; 0 \right\}$ and $M_{t+1}=M_{t}+D(\hat{s}_t, a_t)$, $M_{t+1} = \max \left\{ M_t,\; c(s_t, a_t) \right\}$, and consequently, $\sum_{k=0}^{t} D(\hat{s}_k, a_k)=M_{t+1}=\max_{0 \leq k \leq t} c(s_k, a_k)$. Therefore, constraining the accumulated state-wise cost increments restricts the maximum safety cost encountered during a manipulation trajectory. The original state-wise constraint can consequently be represented by $\mathbb{E}_{\tau \sim \pi}\left[\sum_{t=0}^{\infty} D(\hat{s}_t, a_t)\right]\leq \bar{D}$. This transformation provides a tractable trajectory-level representation of the state-wise safety requirement. Although the two constraints can now be expressed in a trajectory-level form, they retain different safety interpretations. The cumulative constraint regulates the overall amount of unsafe behavior throughout an episode, whereas the state-wise constraint restricts the maximum safety risk encountered during individual interactions.

\subsection{Policy Improvement and Joint Multi-constraint Projection}
Based on the reformulated objective, MDPO adopts a two-stage optimization procedure. At the $k$-th policy iteration, let $\pi_{\theta_k}$ denote the current policy used for trajectory collection. MDPO first performs a reward-oriented policy update to obtain an intermediate policy $\pi_{\theta_{k+\frac{1}{2}}}$. The intermediate policy is subsequently corrected by a single joint safety projection, in which cumulative and state-wise safety signals are simultaneously considered. The overall update is expressed as:
\begin{equation} 
    \label{equ4_1}
    \begin{aligned} 
    \pi_{\theta_k}\!\!\!\xrightarrow{\text{Performance Improvement}}\!\pi_{\theta_{k+\frac{1}{2}}}\!\!\!\xrightarrow{\text{Joint Multi-constraint Projection}}\!\pi_{\theta_{k+1}}.
    \end{aligned}
\end{equation}
The first stage focuses exclusively on improving manipulation performance. Given trajectories collected by the current policy $\pi_{\theta_k}$, we define the policy probability ratio as:
\begin{equation} 
    \label{equ4_2}
    \begin{aligned} 
    x_t(\theta)=\frac{\pi_\theta(a_t \mid \hat{s}_t)}{\pi_{\theta_k}(a_t \mid \hat{s}_t)},
    \end{aligned}
\end{equation}
where $\pi_\theta$ denotes the candidate policy being optimized. Using the reward advantage estimate $\hat{A}^R_t$, MDPO employs the PPO clipped surrogate objective:
\begin{equation} 
    \label{equ4_3}
    \begin{aligned} 
    L_{\mathrm{PI}}(\theta)\!=\!\mathbb{E}_{t}\!\left[\!\min\left(\!x_t(\theta)\hat{A}_t^{R},\operatorname{clip}\left(x_t(\theta),1-\epsilon,1+\epsilon\right)\hat{A}_t^{R}\right)\!\right],
    \end{aligned}
\end{equation}
where $\epsilon$ denotes the clipping coefficient. The intermediate policy is obtained as:
\begin{equation} 
    \label{equ4_4}
    \begin{aligned} 
    \theta_{k+\frac{1}{2}}=\arg\max_{\theta}L_{\mathrm{PI}}(\theta).
    \end{aligned}
\end{equation}
During this stage, cumulative and state-wise safety costs do not directly participate in the actor objective. This separation prevents task reward and heterogeneous safety signals from competing within the same performance improvement update. To prevent excessive policy changes, MDPO additionally monitors the Kullback–Leibler divergence between the current policy and the candidate policy:
\begin{equation} 
    \label{equ4_5}
    \begin{aligned} 
    \bar{D}_{\mathrm{KL}}^{(1)}=\mathbb{E}_{t}\left[D_{\mathrm{KL}}\left(\pi_{\theta_k}(\cdot \mid \hat{s}_t)\Vert\pi_{\theta}(\cdot \mid \hat{s}_t)\right)\right],
    \end{aligned}
\end{equation}
where $D_{KL}$ denotes the KL divergence between the action distributions of two policies. For the Gaussian policy adopted in our implementation, the KL divergence is computed directly from the mean and standard deviation of the two action distributions. If the policy deviation becomes excessively large, the Stage 1 actor update is terminated. After performance improvement, the resulting policy is stored as the intermediate reference policy $\pi_{\theta_{k+\frac{1}{2}}}$.

The intermediate policy obtained from Eq.~\eqref{equ4_4} prioritizes manipulation performance and is therefore not guaranteed to satisfy the safety requirements. MDPO consequently performs a second policy optimization stage that simultaneously corrects cumulative and state-wise safety risks while preserving the intermediate policy as much as possible. Conceptually, the joint projection seeks:
\begin{equation} 
    \label{equ4_6}
    \begin{aligned} \pi_{\theta_{k+1}}\approx\arg\min_{\pi_{\theta}\in\Pi_{\mathrm{cum}}\cap\Pi_{\mathrm{state}}}D_{\mathrm{KL}}\left(\pi_{\theta}\Vert\pi_{\theta_{k+\frac{1}{2}}}\right).
    \end{aligned}
\end{equation}
Rather than independently projecting the intermediate policy with respect to cumulative and state-wise constraints, MDPO incorporates both safety signals into the same projection update. Because the training samples are collected by $\pi_{\theta_k}$, we reuse the importance sampling ratio:
\begin{equation} 
    \label{equ4_7}
    \begin{aligned} 
    x_t^{\mathrm{P}}(\theta)=\frac{\pi_{\theta}(a_t \mid \hat{s}_t)}{\pi_{\theta_k}(a_t \mid \hat{s}_t)},
    \end{aligned}
\end{equation}
and clip it during projection as:
\begin{equation} 
    \label{equ4_8}
    \begin{aligned} 
    \bar{x}_t^{\mathrm{P}}(\theta)=\operatorname{clip}\left(x_t^{\mathrm{P}}(\theta),1-\epsilon_P,1+\epsilon_P\right),
    \end{aligned}
\end{equation}
where $\epsilon_P$ controls the range of importance weights in the safety projection. Using the cumulative-cost advantage $\hat{A}^C_t$, we define $L_C(\theta)=\mathbb{E}_{t}\left[\bar{x}_t^{\mathrm{P}}(\theta)\hat{A}_t^{C}\right]$, and using the state-wise cost increment advantage $\hat{A}^D_t$, $L_D(\theta)=\mathbb{E}_{t}\left[\bar{x}_t^{\mathrm{P}}(\theta)\hat{A}_t^{D}\right]$. Although the two safety objectives are estimated from the same trajectories, their contributions are regulated independently through two adaptive coefficients, denoted by $\omega_C$ and $\omega_D$, for cumulative and state-wise safety, respectively. The joint projection objective is therefore:
\begin{equation} 
    \label{equ4_9}
    \begin{aligned} 
    L_{\mathrm{JP}}(\theta)=\omega_C L_C(\theta)+\omega_D L_D(\theta)+\kappa_k \bar{D}_{\mathrm{KL}}^{(2)}(\theta)
    \end{aligned},
\end{equation}
and the final policy is obtained by:
\begin{equation} 
    \label{equ4_10}
    \begin{aligned} 
    \theta_{k+1}=\arg\min_{\theta}L_{\mathrm{JP}}(\theta).
    \end{aligned}
\end{equation}
Here,
\begin{equation} 
    \label{equ4_11}
    \begin{aligned} 
    \bar{D}_{\mathrm{KL}}^{(2)}(\theta)=\frac{1}{d_a}\mathbb{E}_{t}\left[D_{\mathrm{KL}}\left(\pi_{\theta}(\cdot \mid \hat{s}_t)\Vert\pi_{\theta_{k+\frac{1}{2}}}(\cdot \mid \hat{s}_t)\right)\right],
    \end{aligned}
\end{equation}
where $d_a$ denotes the action dimension and $\kappa_k$ controls the strength of the KL regularization. The role of Eq.~\eqref{equ4_11} differs from that of Eq.~\eqref{equ4_5}. During Stage 1, the KL divergence restricts the performance-oriented policy from deviating excessively from the original behavior policy $\pi_{\theta_k}$. During Stage 2, the KL term instead restricts the safety-corrected policy from moving excessively far away from the intermediate policy $\pi_{\theta_{k+\frac{1}{2}}}$. Therefore, the second-stage optimization jointly decreases cumulative and state-wise safety risks while preserving the task-performance improvement obtained in the first stage.

\begin{figure}[t]
    \centering
    \includegraphics[scale=0.349]{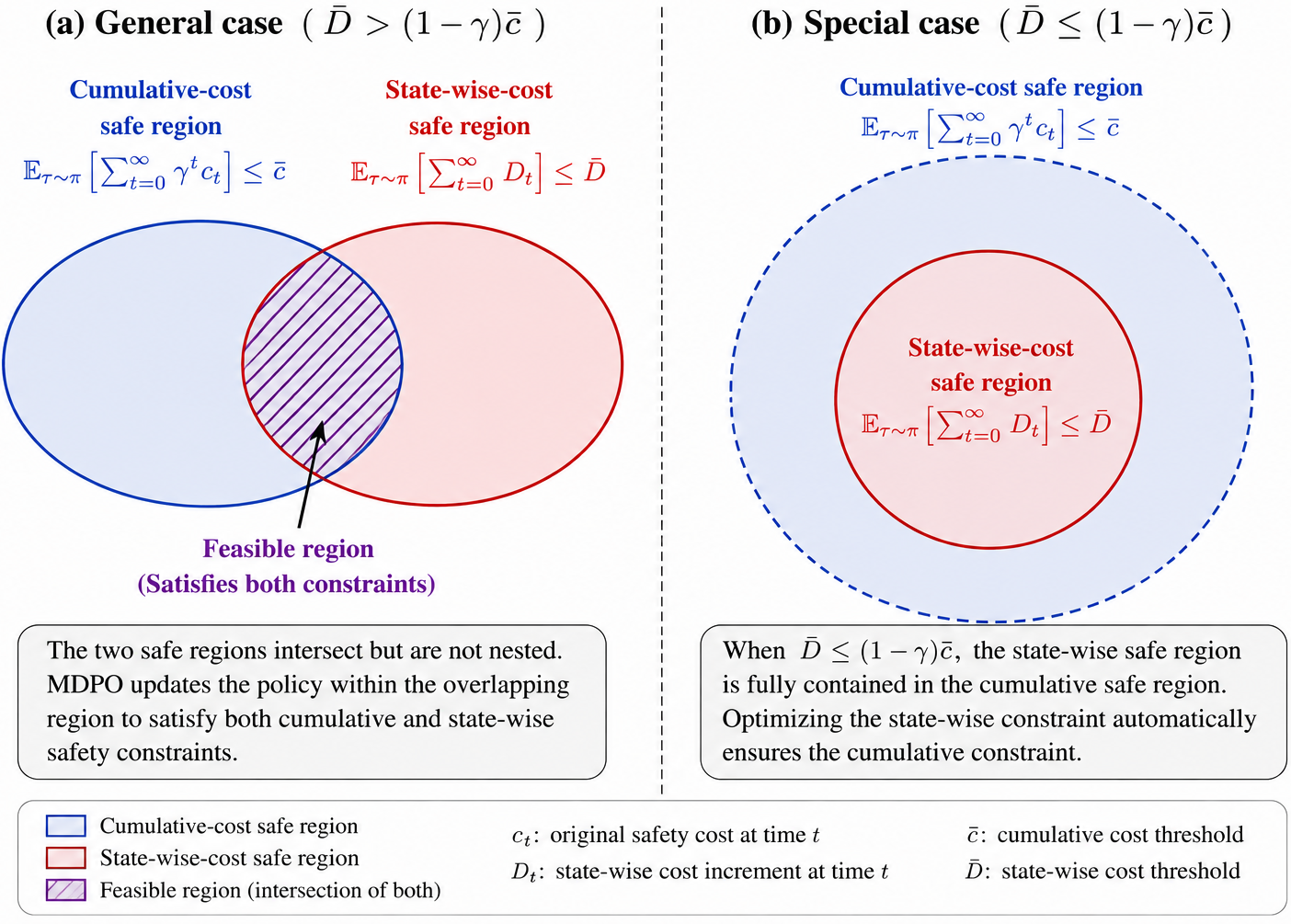}
    \caption{Relationship between the cumulative-cost and state-wise-cost safe regions. (a) In the general case, the two safe regions partially overlap, and their intersection defines the feasible policy region satisfying both constraints. (b) When $\bar{D}\leq(1-\gamma)\bar{c}$, the state-wise-cost safe region is contained within the cumulative-cost safe region, such that satisfying the state-wise constraint also guarantees the cumulative constraint.}
    \label{fig:relationship}
    \vspace{-0.4cm}
\end{figure}

To further clarify the joint projection, we examine the relationship between the feasible regions induced by the cumulative and state-wise safety constraints. In the general case, the two feasible regions are not necessarily nested, since they characterize safety from different temporal perspectives. The cumulative constraint restricts the long-term accumulation of unsafe behaviors, whereas the state-wise constraint limits the maximum safety risk encountered during interaction. Therefore, policies satisfying one constraint do not necessarily satisfy the other, and the feasible policy set considered by MDPO is given by their intersection, $\Pi_{\mathrm{cum}} \cap \Pi_{\mathrm{state}}$, as illustrated in Fig.~\ref{fig:relationship}(a).

A special inclusion relationship arises when the state-wise safety threshold is sufficiently restrictive. Based on the state-wise cost increment formulation, suppose that the transformed state-wise constraint satisfies $\mathbb{E}_{\tau \sim \pi}\left[\sum_{t=0}^{\infty} D_t\right]\leq \bar{D}$, and the corresponding threshold satisfies $\bar{D} \leq (1-\gamma)\bar{c}$. To see the relationship between the two feasible regions, the cumulative cost under policy $\pi$ can be bounded as:
\begin{equation} 
    \label{equ4_12}
    \begin{aligned}
    \mathbb{E}_{\tau \sim \pi}\left[\sum_{t=0}^{\infty}\gamma^t c(\hat{s}_t,a_t)\right] &=\sum_{t=0}^{\infty}\gamma^t\mathbb{E}\left[c(\hat{s}_t,a_t)\right] \\
    &\leq\sum_{t=0}^{\infty}\gamma^t\mathbb{E}_{\tau \sim \pi}\left[\sum_{k=0}^{\infty} D_k\right] \\
    &\leq\sum_{t=0}^{\infty}\gamma^t \bar{D}=\frac{\bar{D}}{1-\gamma}\leq\bar{c}.
\end{aligned}
\end{equation}
The first inequality follows from the state-wise cost increment construction in $\sum_{k=0}^{t} D(\hat{s}_k, a_k)=M_{t+1}=\max_{0 \leq k \leq t} c(\hat{s}_k, a_k)$, since the accumulated increments correspond to the maximum state-wise cost along the trajectory and therefore upper-bound the safety cost at every individual time step. The second inequality follows from the transformed state-wise constraint, while the equality follows from the geometric-series identity. Consequently, any policy satisfying the state-wise safety constraint also satisfies the cumulative cost constraint, which implies $\Pi_{\mathrm{state}} \subseteq \Pi_{\mathrm{cum}}$. Therefore, under $\bar{D} \leq (1-\gamma)\bar{c}$, the joint feasible region reduces to $\Pi_{\mathrm{state}}$, as illustrated in Fig.~\ref{fig:relationship}(b). In the general case where this sufficient condition does not hold, such a containment relationship is not guaranteed, and the two feasible regions may only partially overlap, as shown in Fig.~\ref{fig:relationship}(a). MDPO therefore performs the joint projection toward their intersection to simultaneously account for both safety requirements.

\begin{figure*}[t]
    \centering
    \includegraphics[scale=0.357]{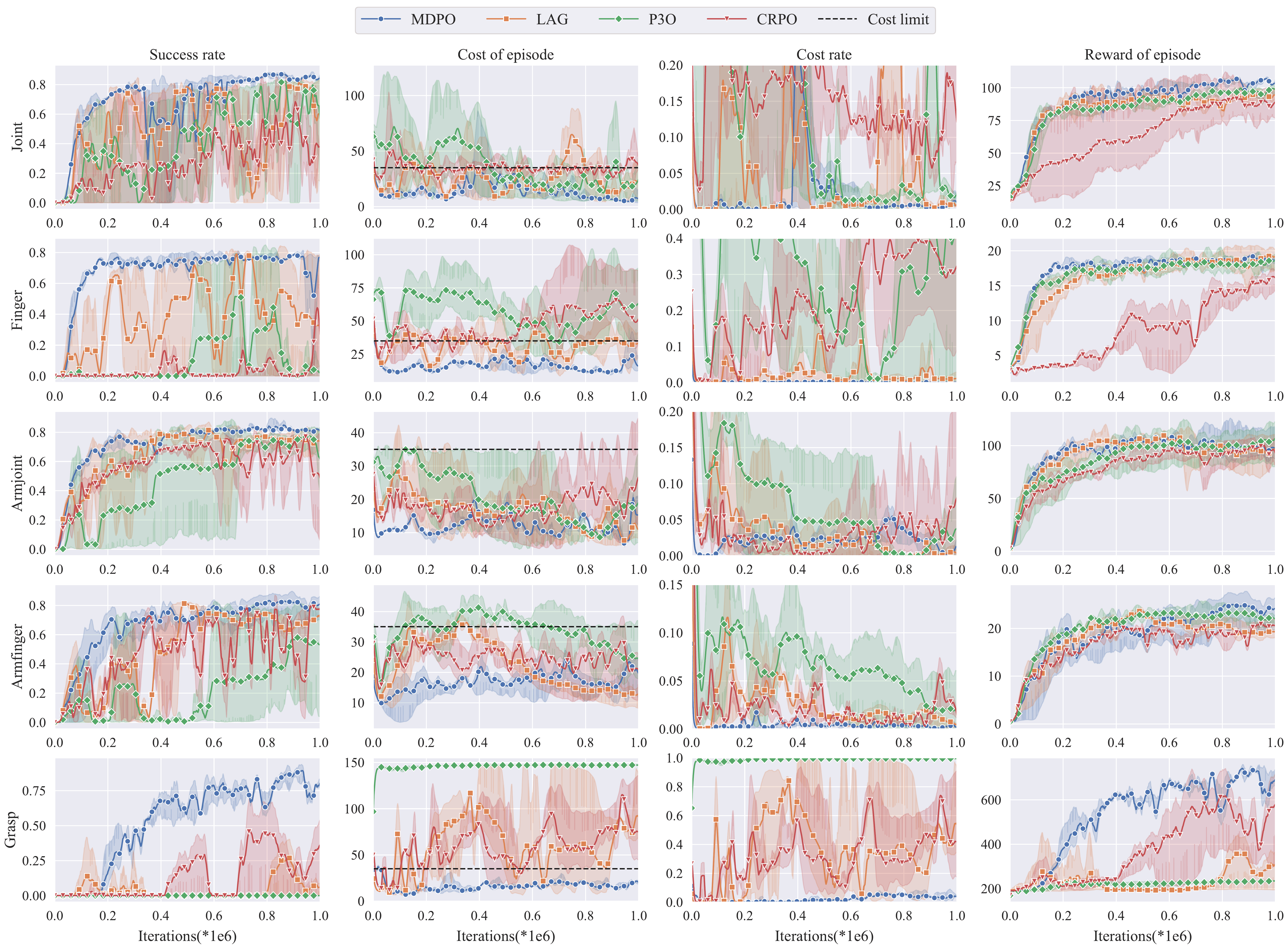}
    \vspace{-0.65cm}
    \caption{Performance comparison of MDPO with LAG, P3O, and CRPO on five dexterous manipulation tasks in ReDMan. From top to bottom, the rows correspond to \textit{Joint}, \textit{Finger}, \textit{Armjoint}, \textit{Armfinger}, and \textit{Grasp}, while the columns report success rate, cost of episode, cost rate, and reward of episode, respectively. The solid curves denote the mean performance over three random seeds, the shaded regions indicate the corresponding standard deviation, and the black dashed lines represent the cumulative cost limits.}
    \label{fig:comparison_allenv}
    \vspace{-0.4cm}
\end{figure*}

\subsection{Practical Policy Optimization}
To implement the two-stage optimization described above, MDPO maintains a reward critic $V_R$, a cumulative-cost critic $V_C$, and a state-wise cost increment critic $V_D$. Given the augmented observation $\hat{s}=(s_t,M_t)$, the corresponding temporal-difference residuals are
\begin{equation} 
    \label{equ4_13}
    \begin{aligned}
    &\delta_t^{R}=r_t+\gamma V_R(\hat{s}_{t+1})-V_R(\hat{s}_t),\\
    &\delta_t^{C}=c_t+\gamma V_C(\hat{s}_{t+1})-V_C(\hat{s}_t),\\
    &\delta_t^{D}=D_t+\gamma V_D(\hat{s}_{t+1})-V_D(\hat{s}_t).
\end{aligned}
\end{equation}
The reward, cumulative-cost, and state-wise cost increment advantages are estimated using generalized advantage estimation (GAE), $\hat{A}_t^{X}=\sum_{l=0}^{T-t-1}\left(\gamma \lambda_{\mathrm{GAE}}\right)^l\delta_{t+l}^{X},\qquad X \in \{R,C,D\},$ where $\lambda_{\mathrm{GAE}}$ denotes the GAE trace parameter. The same $\lambda_{\mathrm{GAE}}$ is used for all three advantage estimators, whereas the contributions of the cumulative and state-wise safety constraints are independently regulated in the subsequent projection stage.

Before performing the joint projection, MDPO estimates the current safety levels from completed trajectories. For cumulative safety, we use the mean episode cost $\hat{C}_k=\frac{1}{N_k}\sum_{e=1}^{N_k}\sum_{t=0}^{T_e-1}c_t^{(e)}$, where $N_k$ is the number of completed episodes at iteration $k$. For state-wise safety, the maximum instantaneous cost is first recorded for each episode, and a high quantile across the completed trajectories is used $\hat{D}_k=Q_q\left(\left\{\max_{t} c_t^{(e)}\right\}_{e=1}^{N_k}\right)$, where $Q_q$ denotes the $q$-quantile. We use $q=0.95$ in our implementation, making the state-wise safety estimate more sensitive to high-risk trajectories rather than allowing severe violations to be diluted by averaging. To reduce fluctuations induced by finite rollout samples, the two safety statistics are independently smoothed using exponential moving averages. Their normalized deviations from the corresponding safety targets are then computed as:
\begin{equation} 
    \label{equ4_14}
    \begin{aligned}
    &v_C=\operatorname{clip}\left(\frac{\hat{C}_k-b_C}{b_C},-1,1\right),\\
    &v_D=\operatorname{clip}\left(\frac{\hat{D}_k-b_D}{b_D},-1,1\right),
\end{aligned}
\end{equation}
where $b_C=\rho_C \bar{c}$ is a conservative cumulative-cost target with $0<\rho_C<1$, and $b_D = \bar{D}$ is the state-wise target. In our implementation, $\rho_C=0.9$. MDPO maintains two independent safety coefficients $z_C$ and $z_D$, which are updated according to their respective constraint deviations:
\begin{equation} 
    \label{equ4_15}
    \begin{aligned}
    z_C\!\!\leftarrow\!\!\Pi_{[0,z_C^{\max}]}\!\left(z_C\!+\!\eta_C h_k v_C\right)\!,\!z_D\!\!\leftarrow\!\!\Pi_{[0,z_D^{\max}]}\!\left(z_D\!+\!\eta_D h_k v_D\right)\!,
\end{aligned}
\end{equation}
where $\eta_C$ and $\eta_D$ are independent adaptation rates, $h_k \in [0,1]$ denotes the gradually activated constraint strength, and $\Pi$ projects the coefficients onto their predefined ranges. Hence, when a safety statistic exceeds its target, the corresponding coefficient increases and imposes a stronger correction; when sufficient safety margin is available, the coefficient is gradually released.

In practice, MDPO further incorporates the recent variation trend of each safety statistic into the effective coefficients. Specifically, a rapidly increasing cumulative or state-wise safety risk strengthens its corresponding correction even before a severe violation is observed. The effective coefficients are additionally attenuated when a sufficiently large safety margin is available, avoiding unnecessary performance degradation. We denote the resulting coefficients used in the joint projection by:
\begin{equation} 
    \label{equ4_16}
    \begin{aligned}
    \omega_C=h_k h_C z_C,\omega_D=h_k h_D z_D,
    \end{aligned}
\end{equation}
where $h_C$ and $h_D$ modulate the correction strength according to the available safety margin. Importantly, $\omega_C$ and $\omega_D$ are independently adapted according to different safety statistics rather than sharing a common weighting coefficient. Using these adaptive safety coefficients, the second-stage joint projection is practically implemented as Eq.~\eqref{equ4_9}.

\section{EXPERIMENTAL EVALUATION}
In this section, we conduct extensive experiments to evaluate the effectiveness of MDPO in multi-constraint safe dexterous manipulation.

\begin{table}[t]
\vspace{+0.1cm}
\centering
\caption{The hyper-parameters of the MDPO algorithm model.}
\vspace{-0.2cm}
\label{table:hyper-parameters}
\scalebox{0.785}{
\begin{tabular}{cccc}
\hline
Hyper-parameters                              & Setting & Hyper-parameters                                & Setting \\ \hline
Discount factor $\gamma$                      & 0.998   & GAE trace parameter $\lambda_{\mathrm{GAE}}$    & 0.95    \\
Clipping coefficient $\epsilon$               & 0.2     & Projection clipping coefficient $\epsilon_P$    & 0.2     \\
Stage-2 KL coefficient $\kappa_0$     & 0.75    & State-wise risk quantile $q$                    & 0.95    \\
Initial cumulative coefficient $z_C^0$ & 1e-4    & Initial state-wise coefficient $z_D^0$   & 1e-5    \\
Cumulative adaptation rate $\eta_C$           & 3e-3    & State-wise adaptation rate $\eta_D$ & 6e-5    \\
Cumulative upper bound $z_C^{max}$            & 0.5     & State-wise upper bound $z_D^{max}$              & 0.02    \\ \hline
\end{tabular}}
\vspace{-0.45cm}
\end{table}

\subsection{Experimental Setup}
\textbf{Tasks.} We evaluate MDPO on the ReDMan~\cite{geng2025redman} benchmark, which provides standardized safety-constrained RL environments for dexterous manipulation. Five representative tasks are considered in our experiments, including \textit{Joint}, \textit{Finger}, \textit{Armjoint}, \textit{Armfinger}, and \textit{Grasp}. These tasks cover different dexterous manipulation scenarios involving hand joints, fingers, coordinated arm-hand control, and object interaction. The original ReDMan environments characterize safety violations mainly using binary cost signals, where the cost changes abruptly between safe and unsafe states. To provide a denser safety signal for policy optimization and better reflect the severity of safety violations, we modify the original cost formulation into a continuous cost function while preserving the corresponding safety semantics of each task. The resulting continuous cost $c_t$ is used to construct both the cumulative and state-wise safety requirements. The running maximum cost $M_t$ is incorporated into the observation, and the state-wise cost increment $D_t$ is constructed to provide a tractable optimization signal for state-wise safety. For all tasks, we set the cumulative cost threshold $\bar{c}=35$ and and state-wise cost threshold $\bar{D}=1$.

\textbf{Baselines.} We compare MDPO with three representative constrained RL methods, including LAG, P3O, and CRPO. Since these methods are originally designed mainly for conventional constrained RL settings, we adapt all of them to the same multi-constraint formulation considered in this work, such that both the cumulative cost constraint and the state-wise cost constraint are explicitly incorporated during policy optimization. Specifically, the multi-constraint LAG baseline introduces independent adaptive multipliers for the cumulative and state-wise safety constraints; P3O is extended by introducing separate penalty terms for the two safety objectives; and CRPO is adapted to account for both constraints when determining whether to optimize task performance or perform safety-oriented policy updates. These adaptations ensure that all baselines operate under the same cumulative and state-wise safety requirements as MDPO. All methods are trained using the same environment interaction budget and evaluated under identical task settings for a fair comparison.

\textbf{Implementation Details.} The hyperparameters involved in the practical implementation of MDPO are summarized in Table~\ref{table:hyper-parameters}. The same settings are used across all tasks.

\subsection{Comparison Experiments}
Fig.~\ref{fig:comparison_allenv} compares MDPO with the multi-constraint baselines LAG, P3O, and CRPO on five dexterous manipulation tasks. Overall, MDPO achieves a favorable balance between task performance and safety. Across the evaluated tasks, it reaches competitive or higher success rates and episode rewards while maintaining relatively low cumulative costs and state-wise violation rates, indicating that the proposed method can effectively coordinate manipulation objectives with heterogeneous safety requirements. From the task-performance perspective, MDPO exhibits faster and more stable convergence on most tasks. On relatively simpler tasks such as \textit{Joint} and \textit{Finger}, several baselines can eventually reach comparable final performance, but generally require more interactions or show larger fluctuations during training. The advantage of MDPO becomes more evident on \textit{ArmJoint}, \textit{ArmFinger}, and particularly \textit{Grasp}, where stronger arm-hand coordination and contact-rich interactions make policy learning more challenging. From the safety perspective, MDPO generally keeps the episode cost below or close to the prescribed cumulative cost limit while simultaneously achieving low cost rates. In contrast, some baselines can satisfy the cumulative-cost requirement but still exhibit notable state-wise violations, indicating that existing constrained RL methods remain insufficient for simultaneously handling heterogeneous safety requirements. This result highlights the advantage of MDPO in jointly regulating long-horizon cumulative risks and instantaneous state-wise safety violations during policy optimization. These results validate the proposed two-stage optimization framework. MDPO first obtains a performance-improved intermediate policy and then jointly corrects cumulative and state-wise safety risks through adaptive projection. The independently adjusted safety coefficients allow the two constraints to contribute according to their corresponding risk levels, while KL regularization limits excessive deviation from the intermediate policy. Consequently, MDPO achieves more stable multi-constraint satisfaction without sacrificing dexterous manipulation performance.

\subsection{Ablation Experiments}
\begin{figure}[t]
    \centering
    \includegraphics[scale=0.331]{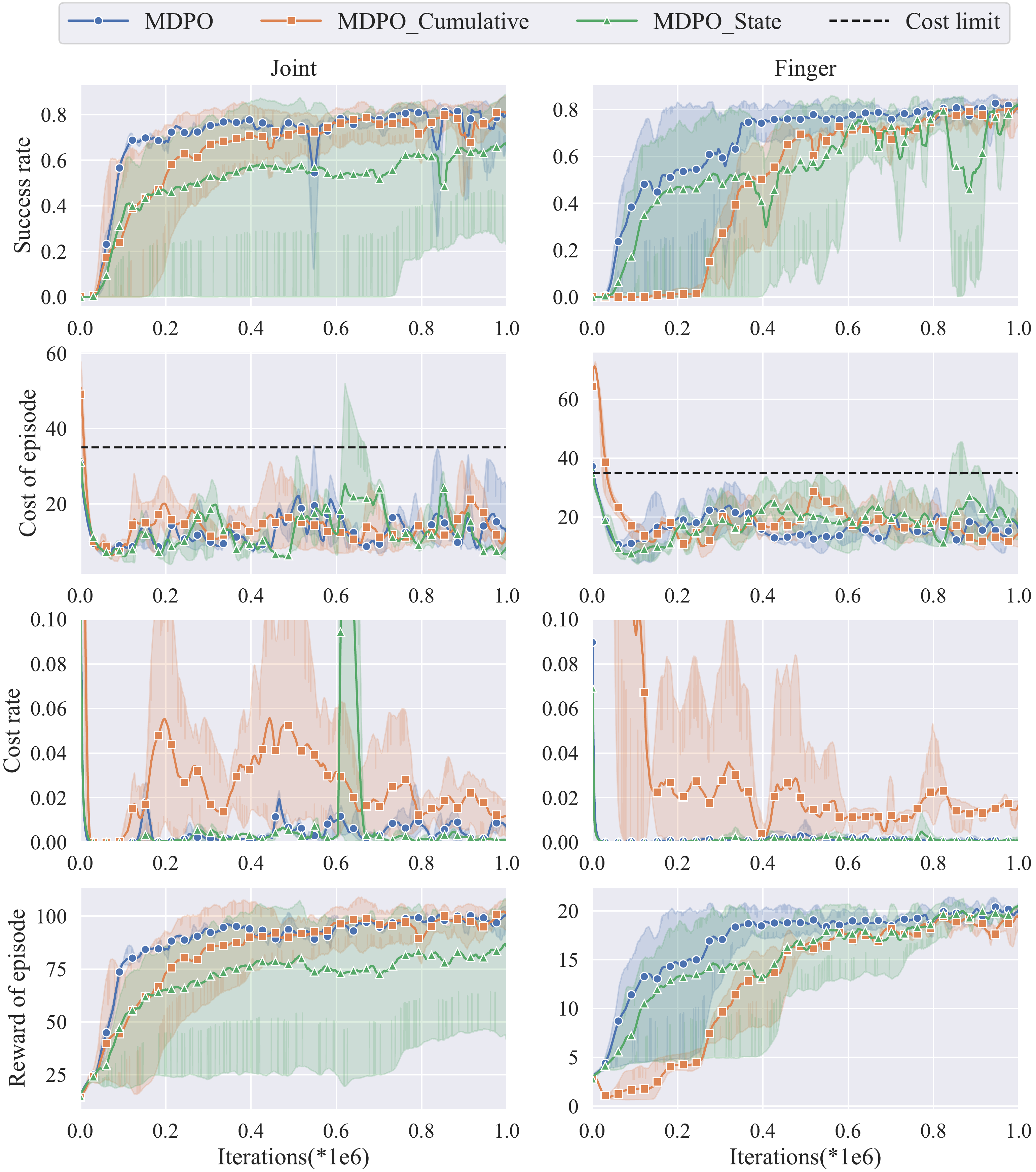}
    \vspace{-0.65cm}
    \caption{Ablation study of MDPO on the \textit{Joint} and \textit{Finger} tasks. MDPO$\_$Cumulative and MDPO$\_$State retain only the cumulative-cost and state-wise safety corrections, respectively. From top to bottom, the rows report success rate, cost of episode, cost rate, and reward of episode.}
    \label{fig:ablation}
    \vspace{-0.4cm}
\end{figure}
Fig.~\ref{fig:ablation} presents the ablation results on the \textit{Joint} and \textit{Finger} tasks. We compare the full MDPO with two single-constraint variants: MDPO$\_$Cumulative, which retains the cumulative-cost safety correction, and MDPO$\_$State, which retains the state-wise safety correction. The comparison is conducted using success rate, cost of episode, cost rate, and reward of episode to investigate the individual and complementary contributions of the two safety constraints. Compared with the full MDPO, MDPO$\_$Cumulative can effectively regulate the episode-level cumulative cost, but generally exhibits a higher state-wise cost rate. This indicates that satisfying the cumulative constraint alone does not necessarily prevent transient unsafe behaviors, since short-duration violations contribute a limited amount to the total episode cost. Explicitly incorporating the state-wise cost signal is therefore important for suppressing safety-critical behaviors at individual interaction steps. MDPO$\_$State effectively reduces state-wise violations, but its task-performance improvement is generally slower and less stable than that of the full MDPO. Relying only on the state-wise constraint does not explicitly regulate the accumulated safety cost over the entire trajectory, which can restrict the balance between long-horizon safety and manipulation performance. In contrast, MDPO jointly exploits cumulative cost and state-wise cost increment advantages, enabling the two safety signals to complement each other during policy projection. Overall, the ablation results demonstrate that neither cumulative nor state-wise safety regulation alone is sufficient to achieve the best performance-safety trade-off. The cumulative constraint controls long-term accumulated violations, while the state-wise constraint suppresses instantaneous safety-critical behaviors. By jointly incorporating both constraints with independently adapted correction strengths, MDPO achieves more stable safety satisfaction while preserving competitive dexterous manipulation performance.

\subsection{Real-world Deployment}
\begin{figure}[t]
    \centering
    \includegraphics[scale=0.309]{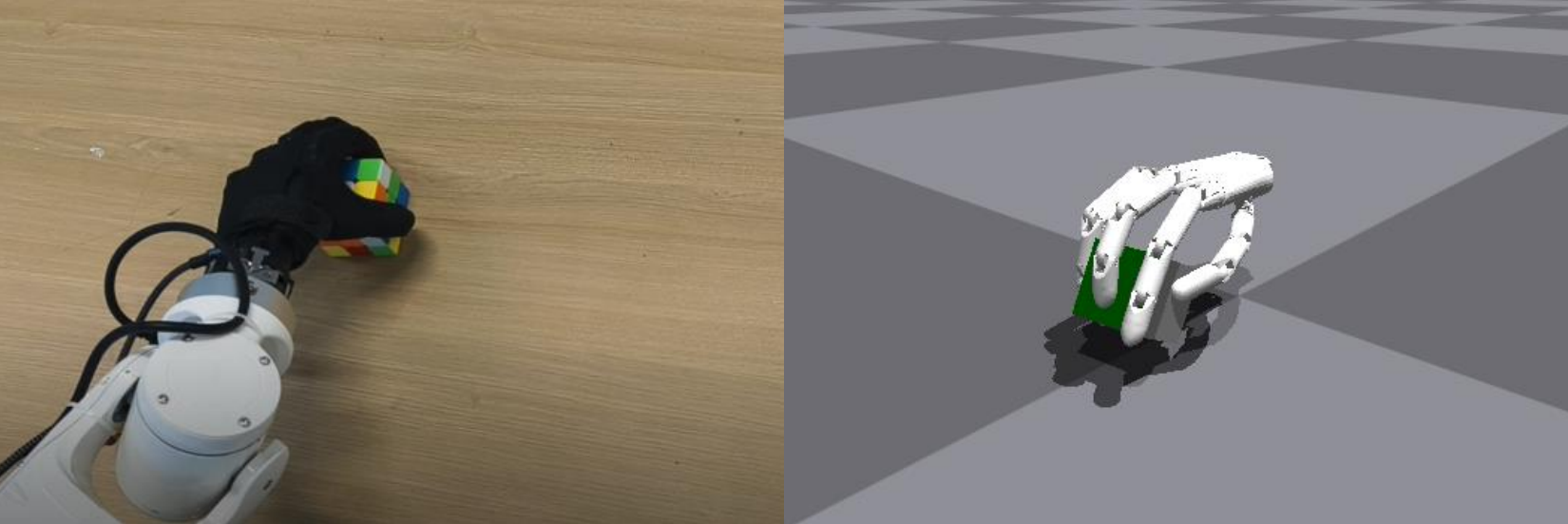}
    \caption{Real-world deployment of MDPO on the \textit{Grasp} task. Left: grasping execution on a Tianji robotic arm equipped with a WujiHand dexterous hand. Right: the corresponding \textit{Grasp} task in simulation.}
    \label{fig:zhenji}
    \vspace{-0.4cm}
\end{figure}

To further evaluate the practical applicability of MDPO, we deploy the learned grasping policy on a real-world robotic platform consisting of a Tianji robotic arm and a WujiHand dexterous hand. As illustrated in Fig.~\ref{fig:zhenji}, we present the corresponding manipulation behaviors in both simulation and the physical environment. The task requires coordinated arm-hand control while maintaining stable object interaction and avoiding aggressive or potentially unsafe hand motions. The policy learned with MDPO can be transferred to the physical platform and successfully execute the grasping behavior while preserving stable and safety-aware manipulation. This result demonstrates the potential of MDPO for transferring multi-constraint dexterous manipulation policies from simulation to real robotic systems.

\section{CONCLUSION}
In this work, we propose the Multi-constraint Dexterous Policy Optimization for safe dexterous manipulation under both cumulative and state-wise safety constraints. To handle the heterogeneous temporal characteristics of these constraints, we reformulate the state-wise requirement into a tractable trajectory-level objective through a state-wise cost increment formulation. Based on this formulation, MDPO adopts a two-stage optimization framework that first improves task performance and then jointly projects the intermediate policy toward the feasible region defined by both safety constraints. The joint projection incorporates cumulative and state-wise safety signals with independently adapted correction coefficients while using KL regularization to preserve the manipulation capability of the intermediate policy. Finally, extensive experiments on ReDMan demonstrate that MDPO achieves a favorable balance between task performance and multi-constraint safety.

\bibliographystyle{IEEEtran}
\bibliography{ref.bib}

\end{document}